%% file: main.tex
\documentclass{article}

\usepackage{iclr2027_conference,times}
\input{math_commands.tex}

\usepackage{amsmath,amssymb,mathtools}
\usepackage{booktabs}
\usepackage{graphicx}
\usepackage{microtype}
\usepackage{xcolor}
\usepackage{url}
\usepackage{hyperref}
\hypersetup{
  hidelinks,
  pdftitle={Role-Decoupled Attention Residuals: Separating Matching and Content Retrieval Across Depth},
  pdfauthor={Kehan Wang}
}
\usepackage{tikz}
\usetikzlibrary{arrows.meta,positioning,fit,backgrounds}

\definecolor{blockblue}{HTML}{3B6FB6}
\definecolor{rdorange}{HTML}{D97928}
\definecolor{softgray}{HTML}{F1F3F5}
\definecolor{inkgray}{HTML}{354052}

\newcommand{\method}{RD-AttnRes}
\newcommand{\parent}{Block AttnRes}
\newcommand{\qk}{\mathrm{QK}}

\title{Role-Decoupled Attention Residuals:\\
Separating Matching and Content Retrieval Across Depth}

\author{Kehan Wang}

\iclrfinalcopy

\begin{document}

\maketitle
\lhead{Preprint.}

\input{sections/00_abstract}
\input{sections/01_introduction}
\input{sections/02_related_work}
\input{sections/03_method}
\input{sections/04_experiments}
\input{sections/05_results}
\input{sections/06_discussion}
\input{sections/07_conclusion}
\input{sections/08_statements}

\bibliography{references}
\bibliographystyle{iclr2027_conference}

\appendix
\input{sections/09_appendix}

\end{document}

%% file: math_commands.tex
\usepackage{amsmath,amsfonts,bm}

\def\eqref#1{equation~\ref{#1}}

\def\1{\bm{1}}

\def\vh{{\bm{h}}}

\def\vq{{\bm{q}}}

\def\vs{{\bm{s}}}

\def\mK{{\bm{K}}}

\def\mQ{{\bm{Q}}}

\def\mV{{\bm{V}}}

\DeclareMathAlphabet{\mathsfit}{\encodingdefault}{\sfdefault}{m}{sl}
\SetMathAlphabet{\mathsfit}{bold}{\encodingdefault}{\sfdefault}{bx}{n}



%% file: sections/00_abstract.tex
\begin{abstract}
Depth-routing residual architectures let a Transformer select among earlier representations instead of inheriting only the immediately preceding state. In Block Attention Residuals, however, one content-dependent depth distribution constructs a shared input for queries, keys, and values. This couples two different jobs: queries and keys determine \emph{where} attention matches, whereas values determine \emph{what} content is retrieved. We ask whether depth selection should respect that distinction. We introduce Role-Decoupled Attention Residuals (\method), a minimal relaxation that shares one route between queries and keys but learns an independent value route over the same residual sources. Tying the two route queries exactly recovers the parent architecture; freeing them adds only one model-width vector per layer and no new token-to-token attention term. Under a frozen, paired pretraining protocol on FineWeb-Edu, five matched seeds at each of 120M and 343M parameters, and a 2.0B-token budget, \method{} improves validation negative log-likelihood in all 10 pairs. Mean reductions are $0.0301$ and $0.0247$, corresponding to mean perplexity reductions of $2.97\%$ and $2.43\%$. Single-seed controls at 0.5B tokens indicate that parameter count, duplicated execution, or a fixed value route does not reproduce the gain, while endpoint routing diagnostics show persistent QK--V divergence. These results support a narrow conclusion: within this training regime, matching and content retrieval benefit from different depth reads. Longer-horizon, cross-data, and stronger-baseline evaluation remains necessary.
\end{abstract}

%% file: sections/01_introduction.tex
\section{Introduction}
\label{sec:intro}

The residual stream is the memory of a Transformer: every layer reads a representation assembled by all layers before it. In the standard architecture, that assembly is fixed by serial addition \citep{he2016resnet,vaswani2017attention}. Recent depth-routing architectures instead let each sublayer retrieve a content-dependent mixture of earlier representations. Attention Residuals (AttnRes) makes this choice explicit, and its memory-efficient Block AttnRes variant routes over block-level sources rather than every layer output \citep{chen2026attnres}. This turns depth from a fixed computational order into a learned retrieval axis.

But learned retrieval introduces a design question that serial residuals never had to answer: \emph{which downstream decisions should share a depth read?} Block AttnRes uses one mixture as the input to all three attention projections. The choice is economical, but it binds together two functionally distinct roles. Queries and keys define a similarity geometry---they determine which token positions match. Values carry the content returned by those matches. A single depth distribution therefore requires the representation that is best for deciding \emph{where to look} also to be best for deciding \emph{what to read}.

We test the smallest possible relaxation of that requirement. Role-Decoupled Attention Residuals (\method) retains the parent source set, block topology, normalization, attention, MLP route, and writeback. It changes only the attention-side depth read: queries and keys share one mixture, while values receive an independent mixture over exactly the same sources (Figure~\ref{fig:method}). This factorization is deliberately asymmetric. Sharing Q and K preserves a coherent matching geometry; separating V frees content retrieval. When the two routing queries are tied, \method{} reduces exactly to Block AttnRes, making the proposal a strict, auditable extension rather than a new residual system with many simultaneous changes.

The empirical question is correspondingly focused: does removing this one coupling improve language-model pretraining when everything else is held fixed? We compare against the parent under matched initialization, data order, optimization, token budget, and validation set. Across 120M- and 343M-parameter models trained for 2.0B tokens, \method{} lowers negative log-likelihood for every one of five preregistered paired seeds at each scale. The effect remains visible at all four evaluation milestones, although its absolute magnitude narrows over training. Preliminary controls and route diagnostics are consistent with the proposed explanation: merely executing two routes or adding parameters does not reproduce the improvement, whereas learned QK and V distributions do not collapse to the same solution.

Our contributions are:
\begin{itemize}
    \item We identify a previously implicit coupling in Block AttnRes: a single depth route jointly determines attention's matching geometry and retrieved content.
    \item We introduce \method, a one-vector-per-layer extension that decouples only the value route, exactly recovers the parent when tied, and adds no new $O(T^2d)$ token-attention operation.
    \item We provide paired evidence at two model scales and five seeds per scale, together with directional controls, route diagnostics, full per-seed results, and an explicit account of backend and evidence limitations.
\end{itemize}

The intended claim is intentionally bounded. We do not establish universal scaling behavior or superiority to every recent residual-routing architecture. We establish that, for the tested same-parent setting, the shared QKV depth read is a consequential constraint rather than an innocuous implementation detail.

%% file: sections/02_related_work.tex
\section{Background and Related Work}
\label{sec:related}

\paragraph{Residual pathways across depth.}
Residual connections stabilize optimization by making identity propagation explicit \citep{he2016resnet}; Pre-Norm Transformers further improve gradient behavior by moving normalization before each sublayer \citep{xiong2020layer}. Several architectures enrich the residual path beyond serial addition. DenseFormer learns static weighted averages of earlier depth states \citep{pagliardini2024denseformer}, while Hyper-Connections and manifold-constrained Hyper-Connections maintain and mix expanded residual streams \citep{zhu2024hyperconnections,xie2025mhc}. These methods broaden communication across depth but do not isolate attention's matching and content roles in the same way as \method.

\paragraph{Content-dependent depth routing.}
AttnRes treats previous residual states as a depth memory and uses a learned pseudo-query to construct token-dependent mixtures \citep{chen2026attnres}. Block AttnRes reduces the source set by grouping layers into blocks and is our direct parent. Concurrent extensions alter other axes of the same framework: Delta Attention Residuals route over incremental rather than cumulative sources \citep{luo2026delta}; Low-Rank Attention Residuals compress the representation used to compute depth scores while retaining full-width values \citep{su2026lowrank}; and Multi-Head Attention Residuals assign depth routes to feature subspaces \citep{luo2026multihead}. In contrast, \method{} keeps the parent's sources and score representation but factorizes its attention-side route by downstream function.

\paragraph{Role-specific cross-layer communication.}
The closest conceptual precedent is MUDDFormer, which learns dynamic dense connections for separate query, key, value, and residual streams \citep{xiao2025muddformer}. Its multiway topology changes how several streams traverse the network. \method{} asks a narrower controlled question inside Block AttnRes: whether QK and V should use the same distribution over an unchanged source set. Depth-Attention also targets cross-layer value access, allowing current queries to attend to keys from previous layers and mix their values inside token attention \citep{zeng2026depthattention}. Value Residual Learning injects earlier-layer values through a fixed cross-layer path \citep{zhou2024valueresidual}. These methods motivate the importance of values, but differ from a pseudo-query depth mixture applied before ordinary causal attention. The distinguishing object in \method{} is not an extra token-attention cache or a fixed value skip; it is a role-factorized read from the parent residual depth memory.

This comparison rules out a broad novelty claim such as ``the first decoupled Q/K/V architecture.'' Our narrower contribution is an exact, minimal factorization of Block AttnRes's shared attention input and a paired test of whether that factorization matters.

%% file: sections/03_method.tex
\section{Role-Decoupled Attention Residuals}
\label{sec:method}

\subsection{The shared-route constraint}

Consider one token position at layer $\ell$ and suppress its batch and sequence indices. Block AttnRes exposes a source set $\mathcal{S}_\ell=\{\vs_i\}_{i=1}^{K_\ell}$ containing the embedding state, completed block outputs, and the current partial block state when available. For learned route query $\vq_\ell\in\mathbb{R}^{d}$, its normalized depth scores and mixture are
\begin{align}
e_{\ell i} &= \vq_\ell^\top \operatorname{RMSNorm}(\vs_i) + b_{\ell i},
&\alpha_{\ell i} &= \frac{\exp(e_{\ell i})}{\sum_{j=1}^{K_\ell}\exp(e_{\ell j})},
&\vh_\ell &= \sum_{i=1}^{K_\ell}\alpha_{\ell i}\vs_i .
\label{eq:parent-route}
\end{align}
The studied implementation uses zero source biases, $b_{\ell i}=0$, and does not divide the route score by $\sqrt d$. One mixture $\vh_\ell$ is normalized and projected to all attention roles:
\begin{equation}
\mQ_\ell=W^Q_\ell\operatorname{RMSNorm}(\vh_\ell),\quad
\mK_\ell=W^K_\ell\operatorname{RMSNorm}(\vh_\ell),\quad
\mV_\ell=W^V_\ell\operatorname{RMSNorm}(\vh_\ell).
\label{eq:shared-qkv}
\end{equation}
Equation~\ref{eq:shared-qkv} is more than parameter sharing: it constrains matching and returned content to be functions of the same convex combination across depth.

\subsection{Factorizing the route by attention role}

\method{} replaces the single attention-side query with $\vq^{\qk}_\ell$ and $\vq^V_\ell$. Each scores the same normalized sources:
\begin{align}
\alpha^{\qk}_{\ell i}
&=\operatorname{softmax}_{i}\!\left((\vq^{\qk}_\ell)^\top\operatorname{RMSNorm}(\vs_i)+b_{\ell i}\right),
&\vh^{\qk}_\ell&=\sum_i\alpha^{\qk}_{\ell i}\vs_i,\\
\alpha^{V}_{\ell i}
&=\operatorname{softmax}_{i}\!\left((\vq^{V}_\ell)^\top\operatorname{RMSNorm}(\vs_i)+b_{\ell i}\right),
&\vh^{V}_\ell&=\sum_i\alpha^{V}_{\ell i}\vs_i.
\label{eq:dual-route}
\end{align}
Queries and keys are projected from $\vh^{\qk}_\ell$, while values are projected from $\vh^{V}_\ell$:
\begin{equation}
\mQ_\ell=W^Q_\ell\operatorname{RMSNorm}(\vh^{\qk}_\ell),\quad
\mK_\ell=W^K_\ell\operatorname{RMSNorm}(\vh^{\qk}_\ell),\quad
\mV_\ell=W^V_\ell\operatorname{RMSNorm}(\vh^{V}_\ell).
\label{eq:rd-qkv}
\end{equation}
Ordinary causal attention then proceeds unchanged. The MLP-side Block AttnRes route, residual source construction, and writeback are also unchanged. Both route queries are zero-initialized, so both distributions begin uniform over the available sources.

\begin{figure}[t]
\centering
\begin{tikzpicture}[
    x=1cm,
    y=1cm,
    font=\footnotesize,
    memory/.style={draw=inkgray!38,fill=white,rounded corners=3pt,
                   minimum width=5.45cm,minimum height=8.0mm,align=center},
    routergray/.style={draw=blockblue!75,fill=blockblue!8,rounded corners=3pt,
                       minimum width=2.70cm,minimum height=7.2mm,align=center},
    routerblue/.style={draw=blockblue!85,fill=blockblue!8,rounded corners=3pt,
                       minimum width=2.28cm,minimum height=7.2mm,align=center},
    routerorange/.style={draw=rdorange!90,fill=rdorange!10,rounded corners=3pt,
                         minimum width=2.12cm,minimum height=7.2mm,align=center},
    stategray/.style={draw=inkgray!48,fill=white,rounded corners=8pt,
                      minimum width=1.50cm,minimum height=5.5mm,align=center},
    stateblue/.style={draw=blockblue!65,fill=white,rounded corners=8pt,
                      minimum width=1.42cm,minimum height=5.5mm,align=center},
    stateorange/.style={draw=rdorange!72,fill=white,rounded corners=8pt,
                        minimum width=1.30cm,minimum height=5.5mm,align=center},
    rolegray/.style={circle,draw=inkgray!60,fill=white,minimum size=7.0mm,
                     inner sep=0pt,font=\bfseries},
    roleblue/.style={circle,draw=blockblue!85,fill=blockblue!9,minimum size=7.0mm,
                     inner sep=0pt,font=\bfseries\color{blockblue}},
    roleorange/.style={circle,draw=rdorange!90,fill=rdorange!10,minimum size=7.0mm,
                       inner sep=0pt,font=\bfseries\color{rdorange}},
    arrgray/.style={-{Latex[length=1.8mm]},line width=0.75pt,draw=inkgray!65},
    arrblue/.style={-{Latex[length=1.8mm]},line width=0.85pt,draw=blockblue},
    arrorange/.style={-{Latex[length=1.8mm]},line width=0.85pt,draw=rdorange}
]
\path[draw=blockblue!40,fill=blockblue!2,rounded corners=5pt,line width=0.8pt]
    (0,0) rectangle (6.55,4.85);
\path[draw=rdorange!52,fill=rdorange!2,rounded corners=5pt,line width=0.9pt]
    (6.85,0) rectangle (13.55,4.85);

\node[anchor=west,font=\bfseries\small,text=inkgray] at (0.34,4.48)
    {\textcolor{blockblue}{A}\quad Block AttnRes};
\node[anchor=east,draw=blockblue!42,fill=blockblue!8,rounded corners=7pt,
      inner xsep=4pt,inner ysep=1.5pt,font=\scriptsize\bfseries,text=blockblue]
      at (6.20,4.48) {SHARED QKV};
\draw[blockblue!20,line width=0.6pt] (0.34,4.16) -- (6.20,4.16);

\node[anchor=west,font=\bfseries\small,text=inkgray] at (7.20,4.48)
    {\textcolor{rdorange}{B}\quad \method};
\node[anchor=east,draw=rdorange!58,fill=rdorange!12,rounded corners=7pt,
      inner xsep=4pt,inner ysep=1.5pt,font=\scriptsize\bfseries,text=rdorange]
      at (13.20,4.48) {OURS};
\draw[rdorange!24,line width=0.6pt] (7.20,4.16) -- (13.20,4.16);

\node[memory] (lmem) at (3.27,3.58)
    {\textbf{Residual depth memory} $\mathcal{S}_\ell$\\[-1pt]
     \scriptsize $\{\vs_1,\vs_2,\ldots,\vs_{K_\ell}\}$};
\node[routergray] (lroute) at (3.27,2.62)
    {shared depth router $\alpha_\ell$};
\node[stategray] (lstate) at (3.27,1.80) {state $\vh_\ell$};
\node[rolegray] (lq) at (2.43,0.90) {$Q$};
\node[rolegray] (lk) at (3.27,0.90) {$K$};
\node[rolegray] (lv) at (4.11,0.90) {$V$};
\draw[arrblue] (lmem) -- (lroute);
\draw[arrblue] (lroute) -- (lstate);
\draw[arrgray] (lstate.south) -- ++(0,-0.23) -| (lq.north);
\draw[arrgray] (lstate.south) -- (lk.north);
\draw[arrgray] (lstate.south) -- ++(0,-0.23) -| (lv.north);
\node[font=\scriptsize,text=inkgray!82] at (2.82,0.28) {\emph{where to look}};
\node[font=\scriptsize,text=inkgray!82] at (4.25,0.28) {\emph{what to read}};

\node[memory] (rmem) at (10.20,3.58)
    {\textbf{Same residual depth memory} $\mathcal{S}_\ell$\\[-1pt]
     \scriptsize $\{\vs_1,\vs_2,\ldots,\vs_{K_\ell}\}$};
\node[routerblue] (rqkroute) at (9.28,2.62)
    {QK router $\alpha_\ell^{\qk}$};
\node[routerorange] (rvroute) at (11.75,2.62)
    {V router $\alpha_\ell^{V}$};
\node[stateblue] (rqkstate) at (9.28,1.80) {$\vh_\ell^{\qk}$};
\node[stateorange] (rvstate) at (11.75,1.80) {$\vh_\ell^{V}$};
\node[roleblue] (rq) at (8.84,0.90) {$Q$};
\node[roleblue] (rk) at (9.72,0.90) {$K$};
\node[roleorange] (rv) at (11.75,0.90) {$V$};
\draw[arrblue] ([xshift=-9.2mm]rmem.south) -- (rqkroute.north);
\draw[arrorange] ([xshift=15.5mm]rmem.south) -- (rvroute.north);
\draw[arrblue] (rqkroute) -- (rqkstate);
\draw[arrorange] (rvroute) -- (rvstate);
\draw[arrblue] (rqkstate.south) -- ++(0,-0.23) -| (rq.north);
\draw[arrblue] (rqkstate.south) -- ++(0,-0.23) -| (rk.north);
\draw[arrorange] (rvstate) -- (rv);
\node[font=\scriptsize,text=blockblue] at (9.28,0.28) {\emph{where to look}};
\node[font=\scriptsize,text=rdorange] at (11.75,0.28) {\emph{what to read}};
\end{tikzpicture}
\caption{The architectural change is role-specific, not source-specific. Both models read the same residual depth memory. Block AttnRes constructs one state for Q, K, and V; \method{} preserves a shared QK matching route (blue) but gives V an independent content route (orange). The only new learned parameter per layer is $\vq_\ell^V$.}
\label{fig:method}
\end{figure}
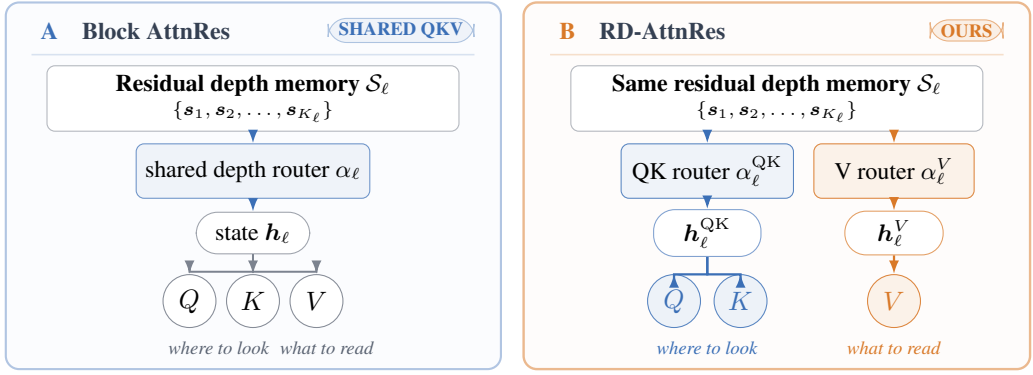

\subsection{Exact recovery and cost}

If $\vq^{V}_\ell=\vq^{\qk}_\ell$ for every layer, then the logits, softmax weights, mixtures, and projected QKV tensors in Equations~\ref{eq:parent-route}--\ref{eq:rd-qkv} are identical to the parent. Thus \method{} contains Block AttnRes as an exactly recoverable submodel; Appendix~\ref{app:recovery} states the argument explicitly.

The only new learned parameters are the independent value queries, one vector in $\mathbb{R}^{d}$ per layer, for $Ld$ total. This is 9,216 parameters (0.0077\%) at 120M scale and 19,968 (0.0058\%) at 343M scale. Computing the second source mixture adds $O(BTK_\ell d)$ work per attention layer but no additional $O(BT^2d)$ token-to-token attention term. This distinction is asymptotic, not a claim of zero runtime cost; measured reference implementations are slower and use more memory (Section~\ref{sec:limitations}).

%% file: sections/04_experiments.tex
\section{Experimental Design}
\label{sec:experiments}

\paragraph{Questions.}
Our design asks four questions. (Q1) Does role decoupling improve validation loss under a paired same-parent comparison? (Q2) Is the effect present throughout the observed training horizon? (Q3) Can parameter count, dual execution, or a fixed value route explain the result? (Q4) Do the learned QK and V routes actually differ?

\paragraph{Data and tokenizer.}
We use the FineWeb-Edu 10BT sample \citep{penedo2024fineweb}. A deterministic salted BLAKE2b-64 document split assigns 99\%/0.5\%/0.5\% of 9,672,101 documents to train/validation/test. The resulting train split contains 9.892B tokens. A 32,768-entry byte-level BPE tokenizer \citep{sennrich2016bpe} is trained only on a deterministic sample of training documents; every document ends with an EOS token. The test split is not used. Main evaluations use a fixed 10,485,760-token validation slice.

\paragraph{Models.}
Both scales are decoder-only Pre-Norm Transformers with RMSNorm \citep{zhang2019rmsnorm}, rotary position embeddings \citep{su2021rope}, SwiGLU MLPs \citep{shazeer2020glu}, tied input/output embeddings, bias-free attention projections, and no dropout. Layers are grouped in blocks of four. The S model has 16 layers, width 576, 36 heads, and 119.79M parameters; the M model has 24 layers, width 832, 52 heads, and 343.04M parameters. Head dimension is 16 and context length is 1,024. Apart from the extra value-route query, parent and \method{} models have identical architecture and initialization.

\paragraph{Optimization and pairing.}
We train with next-token cross-entropy for 61,036 updates, or 2,000,027,648 consumed tokens. Global batch size is 32 sequences (32,768 tokens). AdamW \citep{loshchilov2019adamw} uses $\beta=(0.9,0.95)$, $\epsilon=10^{-8}$, weight decay 0.1 on matrices, and no decay on one-dimensional or routing parameters. Peak learning rates are $3\times10^{-4}$ (S) and $2\times10^{-4}$ (M), with 2\% linear warmup followed by cosine decay to 10\% of peak. Gradients are clipped at 1.0 and training uses BF16 autocast.

For each scale we preregister five seeds: 123, 2025, 2026, 3407, and 7777. Within every pair, Block AttnRes and \method{} use the same initial parent weights, shuffled data order, optimizer schedule, token budget, and evaluation set. We report negative log-likelihood (NLL), perplexity, all paired outcomes, the mean paired difference $\Delta=\mathrm{NLL}_{\mathrm{RD}}-\mathrm{NLL}_{\mathrm{Block}}$, a paired-$t$ 95\% confidence interval, and a deterministic paired-bootstrap interval. With five pairs per scale, intervals describe these seeds and should not be read as broad population guarantees.

\paragraph{Controls and routing diagnostics.}
At 0.5B tokens (seed 123), \emph{Tied} shares the QK/V route, \emph{Average} executes two routes but averages their mixtures, and \emph{Frozen-V} keeps the value route uniform. These are directional controls. At 2B tokens we record the Jensen--Shannon (JS) divergence between QK and V depth distributions at every nontrivial attention layer.

%% file: sections/05_results.tex
\section{Results}
\label{sec:results}

\subsection{Role decoupling wins all paired comparisons}

Table~\ref{tab:main} summarizes the endpoint comparison. \method{} improves NLL in all five paired seeds at both scales. At S, the mean difference is $-0.0301$ (paired-$t$ 95\% CI $[-0.0339,-0.0263]$), corresponding to a 2.97\% mean perplexity reduction. At M, the mean difference is $-0.0247$ (CI $[-0.0396,-0.0098]$) and mean perplexity falls by 2.43\%. The bootstrap intervals, $[-0.0328,-0.0281]$ and $[-0.0347,-0.0163]$, agree in sign. Full per-seed values appear in Appendix~\ref{app:seed-results}.

\begin{table}[t]
\centering
\caption{Validation results after 2.0B training tokens (five paired seeds per scale). NLL entries are mean $\pm$ sample standard deviation across seeds; $\Delta$ is paired RD minus Block. Relative PPL reduction averages the within-seed reductions.}
\label{tab:main}
\small
\begin{tabular}{lrrrrr}
\toprule
Scale & Block NLL & RD NLL & $\Delta$ NLL & PPL $\downarrow$ & Wins \\
\midrule
S (120M) & $3.1724{\pm}0.0102$ & $3.1423{\pm}0.0076$ & $-0.0301{\pm}0.0030$ & 2.97\% & 5/5 \\
M (343M) & $3.0279{\pm}0.0095$ & $3.0032{\pm}0.0073$ & $-0.0247{\pm}0.0120$ & 2.43\% & 5/5 \\
\bottomrule
\end{tabular}
\end{table}

\begin{figure}[t]
\centering
\includegraphics[width=0.64\linewidth]{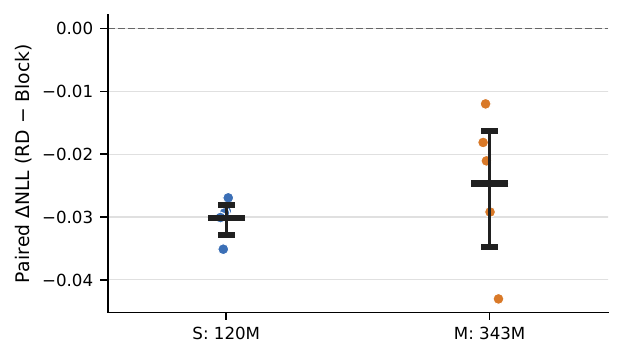}
\caption{Per-seed endpoint effects. Each point is a paired NLL difference; thick bars show means and thin bars show paired-bootstrap 95\% intervals. Negative values favor \method.}
\label{fig:paired}
\end{figure}

The conclusion is not driven by one unusually weak baseline seed. Figure~\ref{fig:paired} shows every paired difference below zero. Variation is larger at M---the gain ranges from 0.0120 to 0.0430 NLL---but its direction remains consistent.

\subsection{The gain persists but narrows over the observed horizon}

\begin{figure}[t]
\centering
\includegraphics[width=\linewidth]{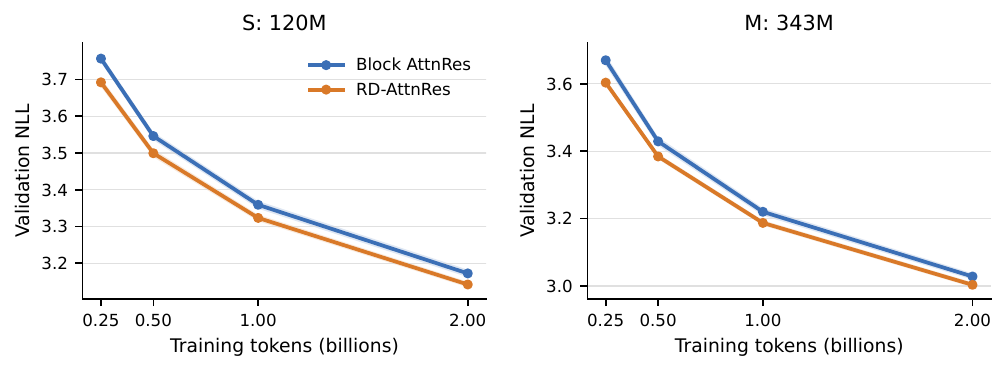}
\caption{Full-validation NLL across the observed training horizon. Lines are means over five paired seeds; bands show $\pm1$ sample standard deviation. Lower is better.}
\label{fig:curves}
\end{figure}

Figure~\ref{fig:curves} aggregates the four full-validation milestones. \method{} is better at 0.25B, 0.5B, 1.0B, and 2.0B tokens at both scales. The mean S gap narrows from $-0.0645$ to $-0.0301$ NLL; the M gap narrows from $-0.0664$ to $-0.0247$. Thus the data support a persistent improvement through 2B tokens, but not a claim that the absolute gap will stay constant or grow under longer training.

\subsection{Directional controls isolate learned role separation}

Table~\ref{tab:ablations} reports the 0.5B-token, seed-123 controls. Tying the routes recovers the parent outcome at S. Executing two routes and averaging them likewise gives no material improvement, arguing against duplicated computation or the extra query vector as sufficient explanations. Freezing V near its uniform initialization also remains close to the parent. Only an independently learned value route yields the large reduction. The same Average and Frozen-V pattern holds at M, where a tied run was not completed. Because these are single-seed controls and the S RD entry comes from a separate ablation run, they narrow explanations but do not establish a universal causal mechanism.

\begin{table}[t]
\centering
\caption{Directional controls at 0.5B tokens (seed 123; validation NLL). The M tied control was not completed.}
\label{tab:ablations}
\small
\begin{tabular}{lccccc}
\toprule
Scale & Block & RD & Tied & Average & Frozen-V \\
\midrule
S (120M) & 3.5416 & \textbf{3.4994} & 3.5413 & 3.5410 & 3.5393 \\
M (343M) & 3.4175 & \textbf{3.3787} & --- & 3.4177 & 3.4175 \\
\bottomrule
\end{tabular}
\end{table}

\subsection{The two routes learn different depth reads}

\begin{figure}[t]
\centering
\includegraphics[width=\linewidth]{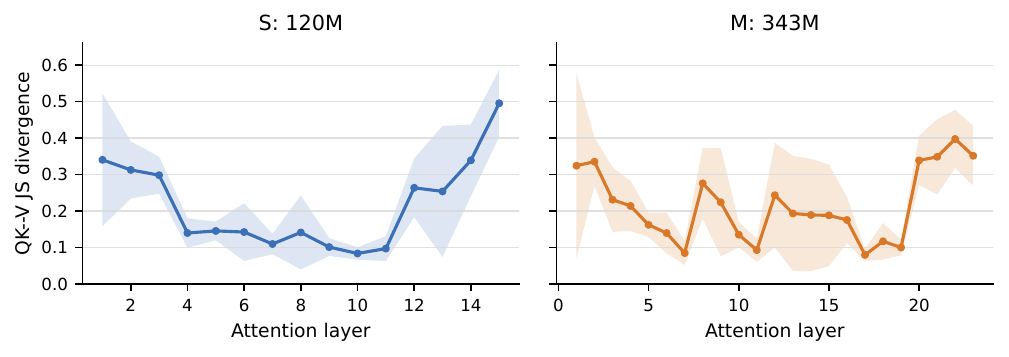}
\caption{Endpoint JS divergence between QK and V depth distributions, by attention layer. Lines average five seeds and bands show $\pm1$ sample standard deviation. Layer 0 is omitted because it has only the embedding source and therefore zero divergence by construction.}
\label{fig:route-js}
\end{figure}

If decoupling merely supplied redundant parameterization, the two distributions could converge to the same route. They do not. Averaging endpoint JS divergence over nontrivial layers gives $0.218\pm0.015$ across S seeds and $0.215\pm0.028$ across M seeds (mean $\pm$ seed-level standard deviation; natural logarithms, maximum $\log 2$). Figure~\ref{fig:route-js} shows that divergence is distributed across depth rather than confined to a single layer. This diagnostic establishes learned differentiation, not that every divergence is useful; the paired loss and controls provide the complementary outcome evidence.

%% file: sections/06_discussion.tex
\section{Discussion and Limitations}
\label{sec:limitations}

\paragraph{What the evidence says.}
The main result supports a specific architectural interpretation. In Block AttnRes, the attention-side depth route is not a neutral shared preprocessing step: constraining matching and content to the same depth distribution measurably worsens short-budget pretraining in the tested models. Exact recoverability, paired runs, controls, and non-collapsed routes make this interpretation more credible than an explanation based only on parameter count.

\paragraph{What remains open.}
The evaluation covers one English web corpus, two sub-billion-parameter models, a 1,024-token context, and 2B training tokens. The shrinking absolute gap makes 4B--10B-token and larger-model runs especially important. We do not report downstream tasks, other domains or languages, long-context behavior, or generation quality. Comparisons with Full AttnRes, Delta AttnRes, Low-Rank AttnRes, Multi-Head AttnRes, MUDDFormer, and Depth-Attention are conceptual rather than matched empirical baselines. An archived Delta baseline was incomplete and is therefore excluded from the result tables. These omissions prevent a current claim of state of the art or broad venue-level completeness.

\paragraph{Systems caveat.}
The extra depth mixture is not free. In seed-123 reference-to-reference measurements, throughput was about 10.1\% lower at S and 8.9\% lower at M, while peak allocated memory increased from 16.01 to 17.65GB and from 21.02 to 23.13GB, respectively. Later fused route kernels improved \method{} throughput, but the corresponding Block runs remained on the reference path, so those measurements are not matched efficiency evidence. Four of five M RD runs and one of five S RD runs use the fused implementation, producing a possible backend/rounding confound. The reference-only seed-123 pair improves at both scales, and all four reference-only S RD pairs improve, but a fully crossed backend study is still required. We therefore claim quality improvements under the archived implementations, not equal efficiency or strict bitwise backend equivalence.

\paragraph{Next decisive tests.}
The most informative additions are: (i) complete matched strong baselines; (ii) repeat the 0.5B controls across seeds; (iii) cross reference and fused kernels within the same seeds; (iv) extend the token horizon; and (v) test whether role separation survives changes in block size, data, scale, and context length. These tests distinguish a general role-factorization principle from a regime-specific optimization benefit.

%% file: sections/07_conclusion.tex
\section{Conclusion}

Selective depth routing should respect the function of what is being routed. Block AttnRes asks one residual mixture to serve both attention matching and content retrieval. \method{} removes only that constraint: Q and K continue to share a route, while V reads the same depth memory independently. The change is exactly reversible by tying the route queries and adds only $Ld$ parameters. Across two model scales and ten paired 2B-token runs, it improves validation loss in every pair; controls and diagnostics are consistent with genuine role specialization. Within the limits of the present study, the result suggests a simple design principle: when depth becomes a learned retrieval axis, sharing should follow downstream role rather than implementation convenience.

%% file: sections/08_statements.tex
\section*{Reproducibility Statement}

Appendix~\ref{app:details} specifies the dataset split, tokenizer, model shapes, optimization, seeds, evaluation budget, parameters, hardware, and software environment. Appendix~\ref{app:seed-results} reports every paired endpoint, Appendix~\ref{app:stats} states the estimands and intervals, and Appendix~\ref{app:backend} inventories implementation backends and reference efficiency measurements. The accompanying Overleaf project includes the exact tabular data used by the figures and a deterministic figure-generation script.

\section*{Ethics Statement}

This work studies language-model architecture on FineWeb-Edu, a filtered web corpus that may retain social biases, personal information, or copyrighted text despite dataset curation. We do not release generated text or make deployment claims. Architectural efficiency and quality improvements can lower experimentation cost, but may also broaden access to capabilities that inherit the risks of web-scale training data. Any downstream deployment requires separate data-governance, safety, and bias evaluation.

\section*{AI Use Statement}

Generative AI tools assisted with literature discovery, manuscript organization and drafting, LaTeX and figure preparation, critical review of the experimental reporting, and interpretation of archived experimental summaries. Experimental runs and logged measurements were produced independently of generative AI. The authors checked AI-assisted claims against source code, raw logs, derived tables, and primary literature, reviewed the complete manuscript, and take responsibility for its contents.

%% file: sections/09_appendix.tex
\section{Complete Paired Results}
\label{app:seed-results}

\begin{table}[ht]
\centering
\caption{All endpoint results after 2,000,027,648 training tokens. Relative PPL reduction is $(\mathrm{PPL}_{\mathrm{Block}}-\mathrm{PPL}_{\mathrm{RD}})/\mathrm{PPL}_{\mathrm{Block}}$.}
\small
\begin{tabular}{lrrrrrr}
\toprule
Scale & Seed & Block NLL & RD NLL & $\Delta$ NLL & Block/RD PPL & PPL $\downarrow$ \\
\midrule
S & 123  & 3.16700 & 3.13796 & $-0.02904$ & 23.736/23.057 & 2.86\% \\
S & 2025 & 3.15800 & 3.13104 & $-0.02695$ & 23.523/22.898 & 2.66\% \\
S & 2026 & 3.17461 & 3.14524 & $-0.02937$ & 23.917/23.225 & 2.89\% \\
S & 3407 & 3.17806 & 3.14798 & $-0.03009$ & 24.000/23.289 & 2.96\% \\
S & 7777 & 3.18425 & 3.14912 & $-0.03513$ & 24.149/23.315 & 3.45\% \\
\midrule
M & 123  & 3.02462 & 3.00649 & $-0.01814$ & 20.586/20.216 & 1.80\% \\
M & 2025 & 3.03657 & 3.00736 & $-0.02921$ & 20.834/20.234 & 2.88\% \\
M & 2026 & 3.03451 & 2.99146 & $-0.04304$ & 20.791/19.915 & 4.21\% \\
M & 3407 & 3.03080 & 3.00973 & $-0.02107$ & 20.714/20.282 & 2.08\% \\
M & 7777 & 3.01297 & 3.00097 & $-0.01200$ & 20.348/20.105 & 1.19\% \\
\bottomrule
\end{tabular}
\end{table}

\section{Exact Recovery of Block AttnRes}
\label{app:recovery}

Assume $\vq^V_\ell=\vq^{\qk}_\ell$ for every layer and that both routes use the same source tensors, RMSNorm, source biases, and softmax. Their logits are then equal elementwise. Softmax preserves equality, so $\alpha^V_{\ell i}=\alpha^{\qk}_{\ell i}$ for every source and $\vh^V_\ell=\vh^{\qk}_\ell$. Substituting into Equation~\ref{eq:rd-qkv} yields Equation~\ref{eq:shared-qkv}. All subsequent attention, MLP routing, and writeback operations are unchanged, hence the complete model is the parent architecture. The Tied control implements this constraint and empirically matches the parent at the observed milestone up to ordinary training nondeterminism.

\section{Full Experimental Details}
\label{app:details}

\subsection{Dataset and tokenizer}

The source archive contains 14 FineWeb-Edu sample/10BT parquet shards totaling 28.518GB and 9,672,101 documents. The deterministic document split contains 9,575,157 train documents (9,892,016,691 BPE tokens), 48,683 validation documents (50,328,166 tokens), and 48,261 test documents (49,144,462 tokens). The tokenizer is trained only on a deterministic one-in-20 sample of training documents: 478,729 documents and 2,274,481,521 UTF-8 bytes. It is byte-level BPE with vocabulary size 32,768 and special IDs PAD=0, UNK=1, BOS=2, EOS=3. Token streams append EOS per document and are stored as unsigned 16-bit integers. The test split was not consulted in method selection or reporting.

\subsection{Model and optimization hyperparameters}

\begin{table}[ht]
\centering
\caption{Model configurations. Parameter counts include tied token embeddings and output head.}
\small
\begin{tabular}{lrr}
\toprule
Configuration & S & M \\
\midrule
Layers / width & 16 / 576 & 24 / 832 \\
Attention heads / head dimension & 36 / 16 & 52 / 16 \\
MLP hidden width & 2,880 & 4,160 \\
Layers per residual block & 4 & 4 \\
Context length / vocabulary & 1,024 / 32,768 & 1,024 / 32,768 \\
Parent parameters & 119,791,296 & 343,039,424 \\
RD parameters & 119,800,512 & 343,059,392 \\
RD increase & 9,216 (0.0077\%) & 19,968 (0.0058\%) \\
Peak learning rate & $3\times10^{-4}$ & $2\times10^{-4}$ \\
Microbatch / accumulation & 8 / 4 & 4 / 8 \\
\bottomrule
\end{tabular}
\end{table}

Both models use RMSNorm $\epsilon=10^{-6}$, RoPE base 10,000, zero dropout, and Gaussian weight initialization with standard deviation 0.02. Attention projections are bias-free; the input embedding and language-model head are tied. Route query vectors are initialized to zero. The global batch is 32 sequences, hence 32,768 tokens per step. Warmup lasts 2\% of 61,036 steps, followed by cosine decay to 10\% of the peak learning rate. Weight decay applies only to matrix parameters. Validation milestones are 250,019,840; 500,006,912; 1,000,013,824; and 2,000,027,648 consumed tokens, each evaluated on 10,485,760 validation tokens.

\subsection{Controls}

The \emph{Tied} control uses the same route query for QK and V. The \emph{Average} control executes the two route operators and averages their mixture outputs before the QKV projections, preserving extra execution and parameterization without role-specific inputs. The \emph{Frozen-V} control leaves the value query fixed at its zero initialization, producing a uniform value distribution while QK learns. All completed controls use seed 123 and stop at 15,259 updates (500,006,912 tokens). The S RD value in Table~\ref{tab:ablations} is an independent ablation-directory run (3.499386), while the corresponding main-run milestone is 3.498424; this small distinction is preserved rather than silently merging runs. The M RD value is the main-run milestone because no separate completed M RD ablation artifact was archived.

\section{Statistical Reporting}
\label{app:stats}

For each seed $s$, the paired effect is
\begin{equation}
\Delta_s=\operatorname{NLL}(\method;s)-\operatorname{NLL}(\parent;s).
\end{equation}
We report $\bar\Delta$, sample standard deviation across the five paired effects, and a two-sided 95\% Student-$t$ interval with four degrees of freedom. A deterministic paired bootstrap resamples the five seed-level differences and uses percentile endpoints. The S mean, standard deviation, $t$ interval, and bootstrap interval are $-0.030118$, $0.003036$, $[-0.033887,-0.026348]$, and $[-0.032828,-0.028064]$. The corresponding M quantities are $-0.024691$, $0.011983$, $[-0.039570,-0.009812]$, and $[-0.034742,-0.016267]$. No independence claim is made beyond the seed-level pairing, and no multiple-comparison correction is needed for the two prespecified scale-level endpoint summaries.

\section{Backend and Efficiency Audit}
\label{app:backend}

Experiments ran as independent single-GPU jobs on two NVIDIA RTX 5090 GPUs (32,607MiB each); no distributed data parallelism was used. The archived environment reports Python 3.12.3, PyTorch 2.12.1+cu130, CUDA 13, cuDNN 92000, Triton 3.7.1, and driver 595.71.05.

All Block AttnRes main runs use the reference route implementation. At S, RD seeds 123, 2025, 2026, and 3407 use the reference implementation and seed 7777 uses the fused Triton path. At M, RD seed 123 uses the reference implementation and seeds 2025, 2026, 3407, and 7777 use the fused path. Both implement the same mathematical operator, but the archive does not contain a complete crossed, tolerance-qualified backend experiment; therefore backend-dependent floating-point differences remain a validity concern.

For seed 123 under reference-to-reference execution, S throughput is 56,158 tokens/s for Block and 50,505 tokens/s for RD (10.1\% lower), with peak allocated memory of 16.005 and 17.647GB. M throughput is 21,616 versus 19,697 tokens/s (8.9\% lower), with peak allocated memory of 21.015 and 23.132GB. Fused RD runs later reach roughly 22.4--22.5k tokens/s at M, but comparison to a reference Block run would conflate method and kernel, so we do not report it as a speedup.

\section{Evidence Boundary and Planned Extensions}

\begin{table}[ht]
\centering
\caption{Claim-to-evidence ledger. This table distinguishes supported conclusions from pending tests.}
\small
\begin{tabular}{p{0.30\linewidth}p{0.30\linewidth}p{0.31\linewidth}}
\toprule
Claim & Current evidence & Boundary \\
\midrule
RD improves the parent at 2B tokens & 10/10 paired wins; two scales; full validation; intervals & One corpus, two small scales, five seeds each \\
The gain is not only parameter count or dual execution & Tied, Average, and Frozen-V controls & Seed 123 at 0.5B; M Tied missing \\
QK and V learn different routes & Nonzero endpoint JS divergence at nearly all nontrivial layers & Diagnostic association, not causal attribution \\
The method is efficient in parameters & Exact $Ld$ overhead & Runtime and memory are measurably higher in reference code \\
The idea generalizes broadly & Not yet established & Needs longer horizon, more scales/data, strong matched baselines \\
\bottomrule
\end{tabular}
\end{table}